\documentclass[journal]{IEEEtran}
\usepackage{times}
\usepackage{soul}
\usepackage{url}
\usepackage[hidelinks]{hyperref}
\usepackage[utf8]{inputenc}
\usepackage{graphicx}
\usepackage{amsmath}
\usepackage{amsthm}
\usepackage{amssymb}
\usepackage{booktabs}
\usepackage{algorithm}
\usepackage{algorithmic}
\usepackage{subfigure}
\usepackage{bbding}
\usepackage{multirow}
\usepackage{xcolor}
\usepackage{wrapfig}
\ifCLASSINFOpdf
\else
\fi
\newcommand{\tabincell}[2]{\begin{tabular}{@{}#1@{}}#2\end{tabular}}

\begin{document}
%
\title{Architecture Aware Latency Constrained Sparse Neural Networks on Mobile Devices}


\author{
  \IEEEauthorblockN{
    Tianli Zhao,
    Qinghao Hu,
    Xiangyu He,
    Weixiang Xu,
    Jiaxing Wang,
    Cong Leng,
    Jian Cheng~\IEEEmembership{Member,~IEEE}
  }
  \IEEEcompsocitemizethanks{
    \IEEEcompsocthanksitem All the authors are with National Laboratory of Pattern Recognition, Institute of Automation, Chinese Academy of Sciences. Bei Jing, China, 100190. (e-mail: zhaotianli2019@ia.ac.cn, huqinghao2014@ia.ac.cn, xiangyu.he@nlpr.ia.ac.cn, xuweixiang2018@ia.ac.cn, wangjiaxing94.gmail.com, lengcong.airia.cn, jcheng@nlpr.ia.ac.cn). \emph{Corresponding Author: Jian Cheng}.
    \IEEEcompsocthanksitem Tianli Zhao, Xiangyu He and Weixiang Xu are also with School of Artificial Intelligence, University of Chinese Academy of Sciences.
    \IEEEcompsocthanksitem This work is under review in IEEE Transactions on Neural Networks and Learning Systems (TNNLS).
  }
}


\markboth{Under Review in IEEE Transactions on Neural Networks and Learning Systems}%
{Zhao \MakeLowercase{\textit{et al.}}: Architecture Aware Latency Constrained Sparse Neural Networks on Mobile Devices}
%



\maketitle


\begin{abstract}

  Acceleration of deep neural networks to meet a specific latency constraint is essential for their deployment on mobile devices. In this paper, we design an architecture aware latency constrained sparse~(ALCS) framework to prune and accelerate CNN models. Taking modern mobile computation architectures into consideration, we propose Single Instruction Multiple Data (SIMD)-structured pruning, along with a novel sparse convolution algorithm for efficient computation. Besides, we propose to estimate the run time of sparse models with piece-wise linear interpolation. The whole latency constrained pruning task is formulated as a constrained optimization problem that can be efficiently solved with Alternating Direction Method of Multipliers (ADMM).   Extensive experiments show that our system-algorithm co-design framework can achieve much better Pareto frontier among network accuracy and latency on resource-constrained mobile devices.
\end{abstract}

%

\section{Introduction}

\IEEEPARstart{M}{ost} recent breakthroughs in artificial intelligence rely on deep neural networks or DNNs as the fundamental building blocks, such as image classification \cite{ALEXNET-2012,VGG-2015,GOOGLENET-2015,INCEPTION-2016,RESNET-2016,MBV1-2017}, object detection \cite{RCNN-2014,FAST-RCNN-2015,FASTER-RCNN-2015,SSD-2016,YOLO-2016}, and so on.
As the emergence of high-end mobile devices in recent years, there is an urgent need to migrate deep learning applications from cloud servers and desktops to these edge devices because of their cost advantages. However, this is challenging due to the high computation intensity of deep learning models and the limited computing power of these devices \cite{ECC-ICLR-2019,ECC-CVPR-2019,HAQ-2019,NET-ADAPT-2018}. In this sense, designing CNN models under a specific latency budget is essential for their deployment on resource-constrained mobile devices. 

There is a considerable body of work in compression and acceleration of deep neural networks to overcome these challenges. Such as network pruning \cite{DEEP-COMPRESSION-2015,LWC-2015,AMC-2018,DMCP-2020,AOWS-2020}, quantization \cite{QCNN-2016,BNN-2016,INTEGER-2018,ADMM-Q-2018}, low rank approximation \cite{ELS-2014,SPEEDUP-LOWRANK-2014,FTCP-2015}, efficient model design \cite{MBV1-2017,EFFICIENT-2019,MBV2-2018,MBV3-2019,OFA-2020,TINYNET-2020,EFFICIENTV2-2021}, and so on. Between which pruning~\cite{DEEP-COMPRESSION-2015,HRANK-2020,DMCP-2020} has been a predominate approach for accelerating deep neural networks. Early endeavours in network pruning often aimed at reducing the model size (e.g. the number of parameters) or the number of Floating Point OPerations (FLOPs) of networks. However, it is recently realized that reducing the number of non-zero parameters or arithmetic operations does not necessarily lead to  acceleration~\cite{EWP-2017,NET-ADAPT-2018}, which is one of the main concerns for model deployment on resource-constrained devices. Resource-constrained compression which aims to directly reduce network latency ~\cite{AMC-2018,AUTOSLIM-2019,AOWS-2020} or energy consumption~\cite{EWP-2017,ECC-CVPR-2019} of networks then emerges and soon draws great attention.

\begin{figure}
    \centering
    \includegraphics[width=0.9\linewidth]{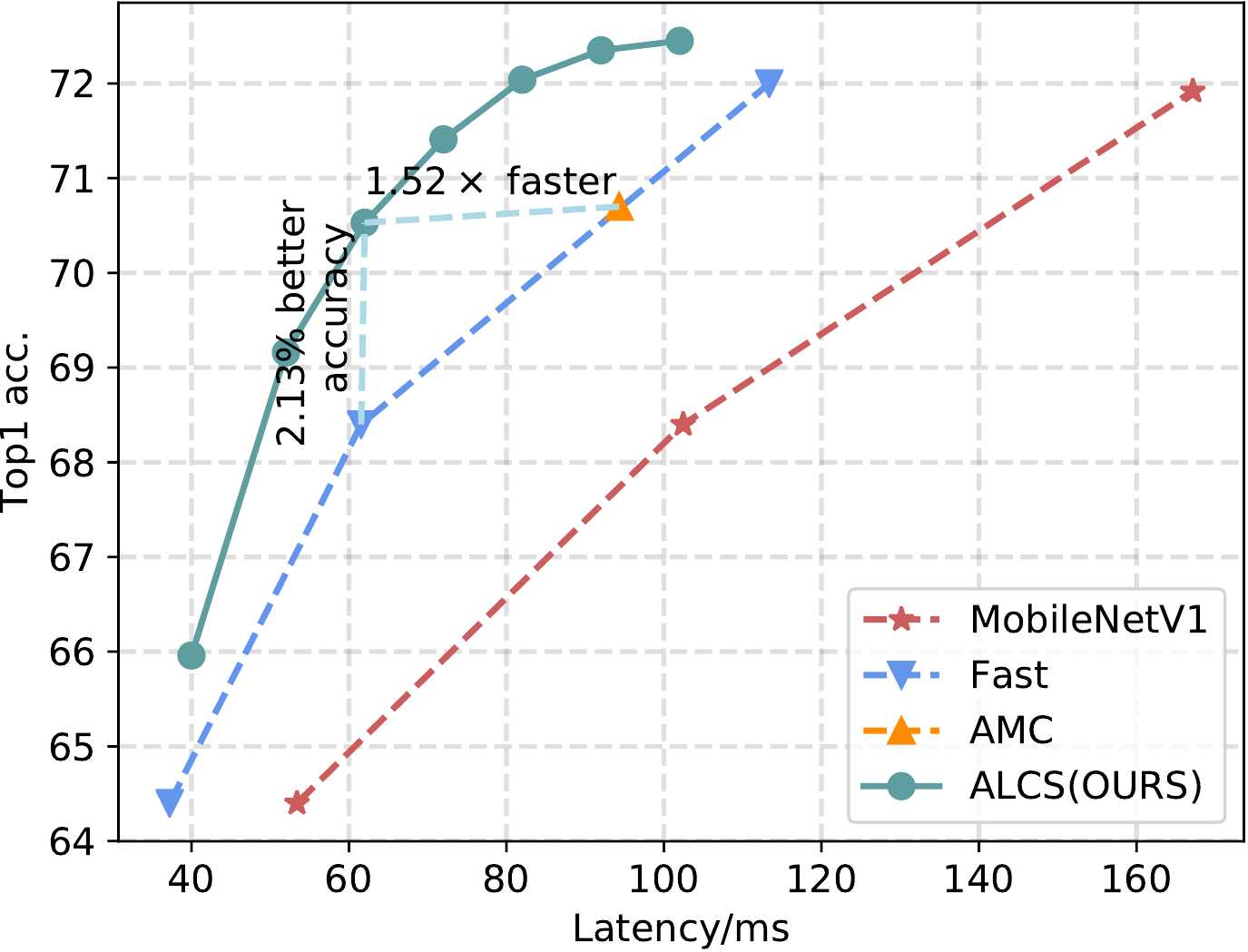}
    \caption{Latency vs. Accuracy for MobileNet \protect\cite{MBV1-2017} on ImageNet. The experiments are conducted on a single ARM Cortex-A72 CPU. Our method outperforms existing acceleration methods such as Fast Sparse ConvNets \protect\cite{FAST-2020} and AMC \protect\cite{AMC-2018}. Best viewed in color.}
    \label{fig:intro-compare}
\end{figure}

While achieving good trade-off among accuracy and latency/energy, there is still space to further push the frontier of resource-constrained compression.

The way is to take advantage of modern mobile computation architectures. The pruning patterns are mostly not specifically designed for mobile devices, pruning is conducted either channel-wisely \cite{EWP-2017,ECC-CVPR-2019,AUTOSLIM-2019,AOWS-2020}, which is too inflexible to attain high compression rate, or randomly~\cite{DEEP-COMPRESSION-2015,LWC-2015}, which is not convenient for acceleration because of their irregular memory access. It is necessary to take into account the computing architectures and design specialized pruning patterns to further push the frontier among network accuracy and latency.

It requires efficient and accurate estimation of latency/energy for solving the constrained problem.
Latency/energy modeling in previous works~ \cite{GSL-2017,EWP-2017,ECC-ICLR-2019} are tied to specific hardware platforms, and the estimation requires deep knowledge to the hardware platform. Other platform independent methods approximate the latency/energy with a look up table \cite{NET-ADAPT-2018,HAQ-2019} or an estimation model~ \cite{ECC-CVPR-2019,AOWS-2020}. However, constructing the look up table or training the estimation model require a large amount of sparsity-latency/energy pairs, which is laborious to collect. 

In this paper, we propose architecture-aware latency constrained sparse neural networks~(ALCS) towards better Pareto frontier among network accuracy and latency. Specifically,  considering that most modern mobile devices utilize the Single Instruction Multiple Data (SIMD) technique to improve the computation capacity,  we propose SIMD-structured pruning which groups the parameters according to the bit-length of SIMD registers. Parameters are then pruned in a group-wise manner. Along with it, we also propose an efficient computation algorithm for accelerating the SIMD-structured sparse neural networks. Our method does not suffer from strong structure constraint as channel pruning and therefore is able to achieve relatively high compression/acceleration ratio on mobile devices. 

For efficient latency estimation, we approximate with piece-wise linear interpolation. Our construction of the latency estimator doesn't require any specific architecture knowledge. Compared to other platform independent estimation methods~\cite{NET-ADAPT-2018,ECC-CVPR-2019,AOWS-2020} which requires tens of thousands of  sparsity-latency pairs, our piece-wise linear interpolation latency estimator is much easier to establish. Only a small collective of  sparsity-latency data pairs~(11 in our experiments) are required.

The whole latency constrained sparsify task is formulated as a constrained optimization problem, which can be efficiently solved with Alternative Direction Method of Multipliers (ADMM). Extensive experiments show that ALCS can achieve better Pareto frontier among network accuracy and latency as shown in Figure \ref{fig:intro-compare}. With ALCS, the execution time of MobileNet is reduced by $2.04\times$ without accuracy drop. The latency of ResNet-50 can be reduced by $1.67\times$ even with $0.11\%$ accuracy improvement. 

In summery, our contributions are three-folds:
\begin{itemize}
    \item {We propose ALCS, an end-to-end system-algorithm co-design framework which utilizes SIMD-structured pruning to exploit modern mobile computation architectures for agile and accurate model.}
    \item {We propose an efficient piece-wise linear interpolation method to estimate the network inference latency, which is sample efficient and accurate.}
    \item {Extensive experiments and ablation studies demonstrate the advantages of architecture-aware pruning, as well as the superiority of our proposed method against a set of competitive compression and acceleration methods.}
\end{itemize}
\section{Related Works} 
\noindent\textbf{Network Pruning.} Network pruning is a key technique for compression and acceleration of neural networks. Pioneer approaches prune weights of models randomly, which means that each individual element of the parameters can be removed or retained without any constraint. This category of pruning method can be dated back to Optimal Brain Damage (OBD) \cite{OBD-1989}. OBD prunes weights based on Hessian matrix of the loss function, which is difficult to get when the amount of parameters becomes large. More recently, Han et al. present a 'Deep Compression' pipeline \cite{DEEP-COMPRESSION-2015}, which prunes parameters with relatively small magnitude. Ding et al. \cite{MOMENTUM-2019} utilize the momentum term of SGD step to force the parameters to converge to zeros. Besides, there are many other works focusing on training a pruned network from scratch \cite{SET-2018,DEEPR-2019,SM-2019,SR-2019,DPF-2020}. These methods can remove a large part of parameters with negligible accuracy loss while are not convenient for inference acceleration because of their irregular memory access \cite{SSL-2016}. 

The limitations of random weight pruning described above motivate recent works \cite{SSL-2016,SFP-2019,DMCP-2020,HRANK-2020,DHP-2020,APS-2020} to focus more on channel pruning, which prunes the parameters in a channel-wise manner. The channel pruning is able to accelerate the computation of networks, while it requires to prune a whole channel simultaneously, which is too inflexible to achieve high compression and acceleration ratio. Moreover, these methods often aim to reduce the model size of networks, while it has been well acknowledged now that network latency, which is one of the main concerns when deploying CNNs on resource-constrained mobile devices, does not decrease monotonously with the reduction of model size \cite{NET-ADAPT-2018}.

\noindent\textbf{Resource Constrained Compression.} Recognizing that model size is not a sufficient surrogate for network latency/energy consumption, recent researchers have started investigating resource constrained compression which compress the network to meet some budgets (e.g. latency, energy). Given some explicit resource constraints, these methods search for the optimal network structures with reinforcement learning \cite{AMC-2018}, greedy search \cite{EWP-2017,NET-ADAPT-2018,AUTOSLIM-2019}, bayesian optimization \cite{CAP-2018}, dynamic programming \cite{AOWS-2020}, or optimize the network structures and values of weights simultaneously with optimization algorithms \cite{ECC-ICLR-2019,ECC-CVPR-2019,DSA-2020}. Compared to previous works, our work further takes into account the computing architecture of mobile devices and propose mobile-oriented SIMD-structured pruning for CNNs. What's more, we employ linear interpolation for estimation of network latency, which is efficient and accurate and needs neither deep architecture knowledge nor large number of collective sparsity-latency data pairs.

\noindent\textbf{Efficient Sparse Computation.} Recently, \cite{FAST-2020} propose efficient computing algorithm for sparse matrix multiplication on mobile devices. The common points between our works is that channel pruning is not necessary for network acceleration on mobile devices, while our method is different from theirs in that our work supports not only matrix multiplication but also general convolution, and we further argue that SIMD-structured pruning is necessary to achieve better trade-off between network accuracy and latency. 
\section{Methodology}

\subsection{Problem formulation}
Our goal is to accelerate the network to meet some latency budget while minimizing the target loss:
\begin{equation}
    \begin{split}
        \min_{W} ~~ & \mathcal{L}(W) \\
        s.t.     ~~ & \mathcal{T}(W) \le T_{budget}
    \end{split}
    \label{eq:problem-def}
\end{equation}
where $W=\{W^{(l)}\}_{l=1}^{L}$ denotes the set of parameters of each layer, $\mathcal{L}(W)$ is the task-specified loss function, for example, the cross entropy loss function for classification. $\mathcal{T}(W)$ and $T_{budget}$ denote the latency of the network and the target latency budget, respectively. There are three important challenges to obstacle for solving the above problem: 1) how to utilize modern computation architectures to get higher compression and acceleration rate on mobile devices, 2) how to efficiently estimate the latency $\mathcal{T}(W)$ of the network, and 3) how to solve the constrained optimization problem. In this work, we propose SIMD-structured pruning along with an efficient SIMD-structured sparse convolution algorithm for CNN acceleration. The network latency is estimated with piece-wise linear interpolation and the constrained problem is finally solved with ADMM. We will introduce more details in the following sections.

\subsection{SIMD-structured pruning for fast inference}
\label{sec:simd-structured-pruning}


It is necessary to take into account the computing architectures of the target platforms for fast inference of CNNs. To this end, considering that most mobile CPUs utilize the Single Instruction Multiple Data (SIMD) technique to improve the computation efficiency, we propose SIMD-structured pruning along with an efficient SIMD-structured sparse convolution algorithm for CNN acceleration. We will describe them in detail in the following sections.

\begin{figure}[!bt]
    \begin{center}
        \includegraphics[width=\linewidth]{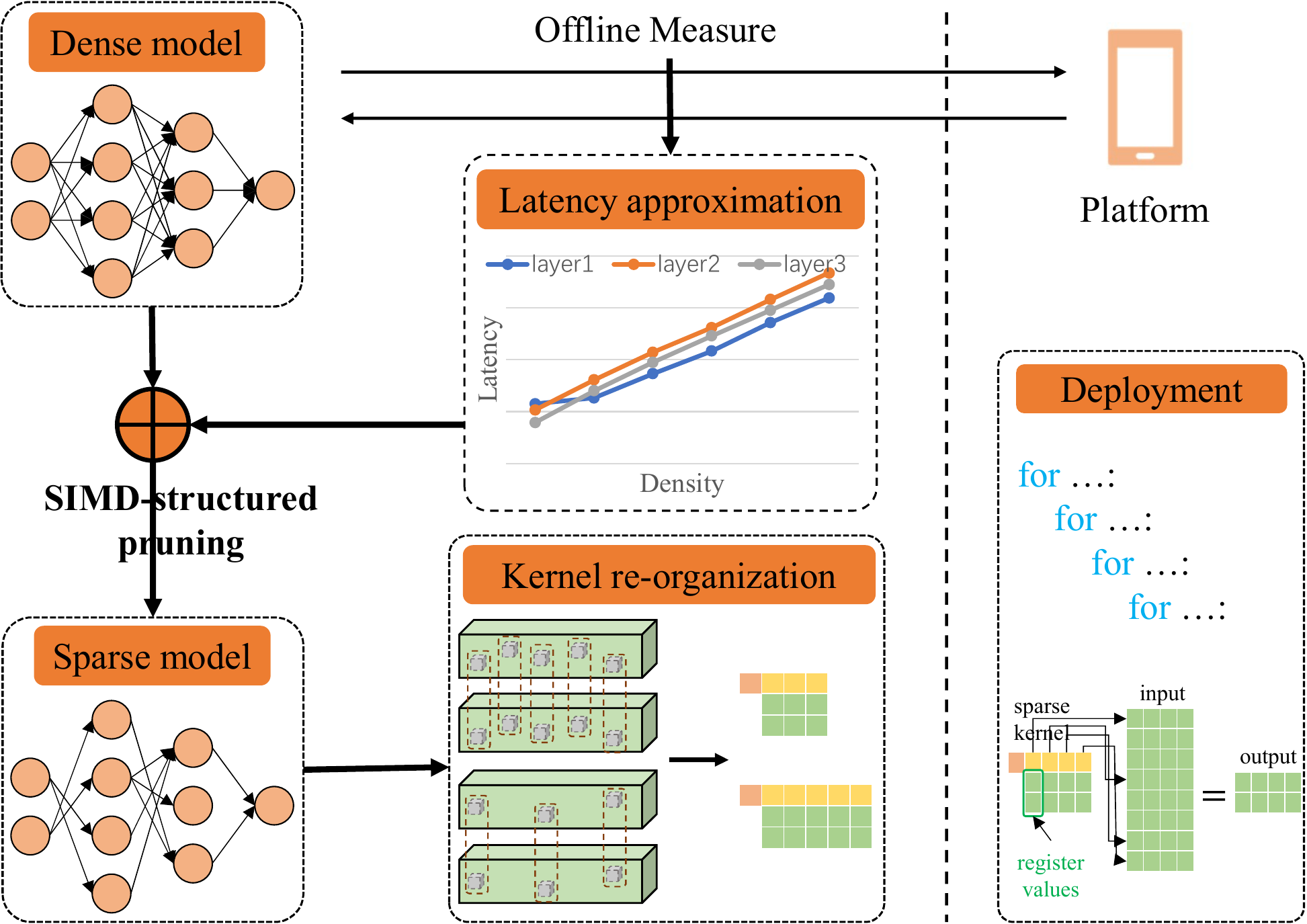}
    \end{center}
    \caption{Framework of architecture aware latency constrained sparse neural networks. \textbf{Left:} In ALCS, network is pruned and accelerated with SIMD-structured pruning, in which parameters are grouped according to the bit-length of SIMD registers in hardware, and pruned in a group-wise manner. \textbf{Middle:} To solve the latency constrained problem, we approximate the latency of compressed models precisely and efficiently with piece-wise linear interpolation. \textbf{Right:} After training, the compressed models can be deployed on the target platform for practical applications.}
    \label{fig:overview}
\end{figure}

\subsubsection{SIMD-structured pruning}
\begin{figure}[!bt]
    \centering
    \includegraphics[width=\linewidth]{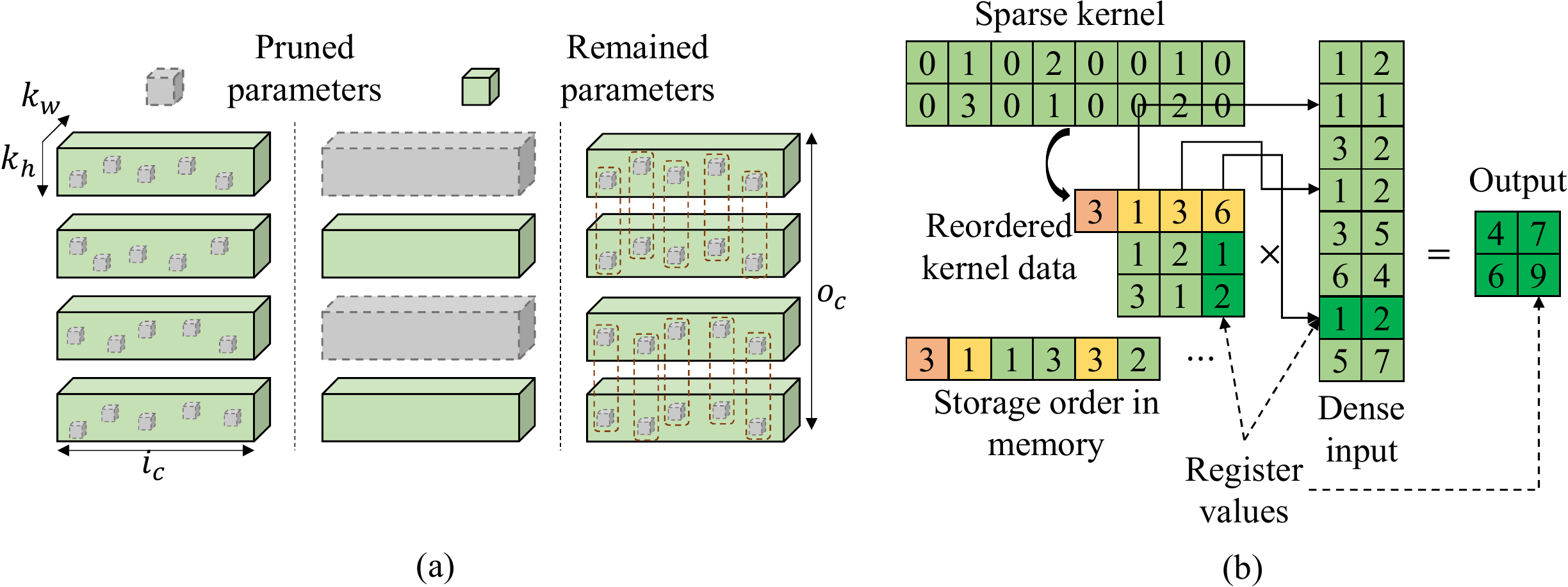}
    \caption{(a) Comparison of \textbf{left}: non-structured pruning method, \textbf{middle}: structured channel pruning, and \textbf{right}: our proposed SIMD structured pruning. In SIMD structured pruning, parameters are divided into groups according to the bit-length of SIMD registers and removed in a group wise way. (b) The storage format of sparse kernel and the core computation component of the proposed SIMD structured sparse convolution algorithm.}
    \label{fig:SIMD-pruning}
\end{figure}
In this section, we introduce the proposed SIMD-structured pruning. Before we get into more details, it is worthy to have a brief introduction to Single Instruction Multiple Data (SIMD). As a data level parallelism scheme, SIMD is widely used in modern mobile CPU architectures. It allows CPUs to operate on a  set of data items at the same time in a single instruction. In this way, a vector of data can be loaded into vector registers and processed simultaneously by one instruction.

We start with grouping of parameters. For a convolutional layer $l$ with filters $W\in \mathbb{R}^{o_c\times i_c\times k_h\times k_w}$, where $o_c, i_c, k_h, k_w$ denote the number of output/input channels and kernel size, respectively. 
The elements are first grouped along the $o_c$ dimension. The size of each group depends on the length of SIMD vector registers. For example, on the widely used ARM v7/v8 architectures, the length of each vector register for SIMD instructions is 128 bits, so for 32-bit single float precision parameters, the group size should be 4. In the other words, parameters at the same location of each 4 adjacent channels are grouped together. The parameters are then pruned in a group wise manner. The right of Figure \ref{fig:SIMD-pruning}(a) shows a simple example for the proposed SIMD-structured pruning with group size of 2. Note that the only constraint of SIMD structured pruning is that the locations of zeros in each group of filters should be the same, however, the locations of zeros across different groups of filters can be irregular. 

\subsubsection{SIMD-structured sparse convolution}
Having introduced the SIMD-structured pruning for deep neural networks, in this section, we describe the efficient algorithm for computation of sparse convolutions. 

\begin{figure}
  \centering
  \includegraphics[width=0.9\linewidth]{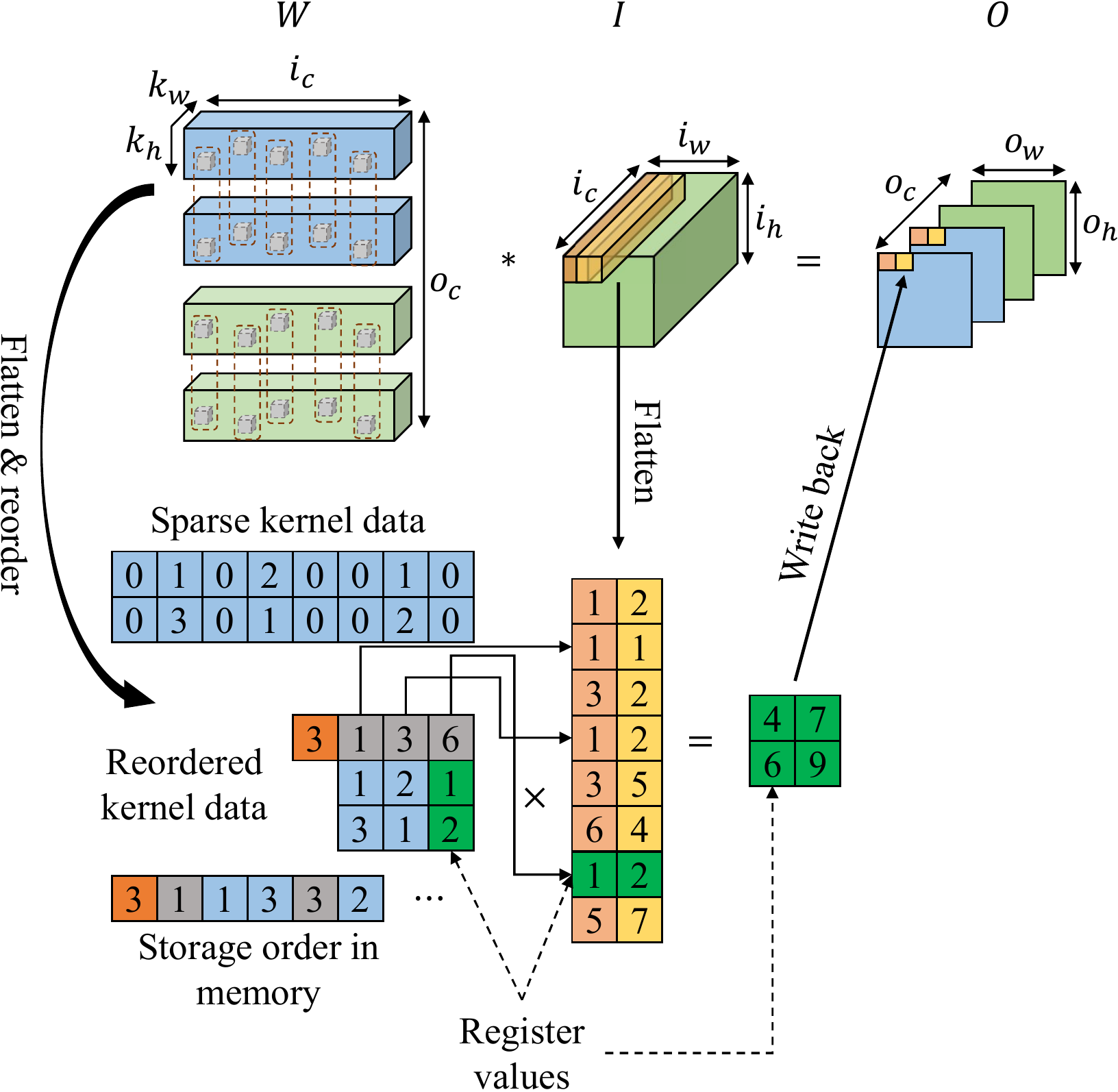}
  \caption{SIMD structured pruning and the efficient sparse convolution algorithm. In this figure, $W, I, O$ denote the kernel, input and output tensors of convolution. The size of input/output feature maps are denoted by $i_c, i_h, i_w (o_c, o_h, o_w)$, respectively, and the kernel size are denoted by $k_h, k_w$. }
  \label{fig:spc-detail}
\end{figure}

\noindent{\textbf{OVERVIEW}} We show in Figure \ref{fig:spc-detail} an overview of the proposed computation algorithm. In this example, the group size of SIMD-structured pruning is 2. We denote the input/output of convolution as $I/O$. The output element at location $(h, w)$ of the $n^{th}$ channel can be computed by:
\begin{equation*}
    O(n, h, w) = \sum_{c=1}^{i_c}\sum_{r=0}^{k_h-1}\sum_{s=0}^{k_w-1} W(n, c, r, s)I(c, h + r, w + s)
\end{equation*}
which can be treated as the inner product of the stretched $n^{th}$ kernel of $W$ and a sub-tensor of $I$ related to the spatial location $(h, w)$. For instance, the element at the top left corner of the first channel of $O$ can be computed by the inner product of the stretched first filter of $W$ and the sub-tensor colored in orange at the top left corner of $I$. It is easy to realize that the output values related to multiple output channels and spatial locations can be computed collectively. For instance, the elements related to the first 2 channels and the first 2 spatial locations of the output $O$ can be computed by: First flattening and stacking together the first 2 filters of $W$ into rows of a matrix, say $W^*$; Then vectorizing and stacking together the sub-tensors of $I$ related to the first 2 output spatial locations (e.g. the sub-tensors colored in orange and yellow, respectively located at the top-left corner of $I$) into different columns of a matrix, say $I^*$; Finally, the 4 output elements can be computed by multiplication between $W^*$ and $I^*$. In case where $W$ is SIMD-structured sparse, this data parallelism can be easily achieved with the help of SIMD instructions. Before going into more details, we first introduce the storage format of SIMD-structured sparse tensor $W$.

\noindent\textbf{STORAGE FORMAT OF SIMD-STRUCTURED SPARSE TENSOR.} As shown in the middle of Figure \ref{fig:spc-detail}, a group of filters is stored in memory in a grouped version of Compressed Sparse Row (CSR) format, which consists of the number of non-zero columns (the orange value), column offsets (the gray elements), and non-zero values (the light blue elements). For instance, we can see in middle of Figure \ref{fig:spc-detail} that, there are 3 non-zero columns in the original kernel data, so the orange value is 3. For the first non-zero column $[1, 3]^T$, there is 1 column before it in the original kernel data, so the column offset related to $[1, 3]^T$ is 1. As for the second non-zero column $[2, 1]^T$, there are 3 columns before it in the original kernel data, so its column offset is 3, and so on. Note that after training, the values of parameters are fixed during inference, so this re-organization of parameters can be done in advance without any extra time overhead.

\noindent\textbf{EFFICIENT MULTIPLICATION COMPUTATION.} As what has been described in the previous Section, the core computation component is the multiplication between the SIMD-structured sparse matrix $W^*$ and the dense input $I^*$. In this section we describe how this multiplication can be efficiently computed when $W^*$ is stored in the format as described in the previous section. See the middle of Figure \ref{fig:spc-detail} as a simple example. We first allocate several SIMD registers and initialize them to zero for storage of intermediate results. Then we load from memory the number of non-zero columns of the kernel data, which determines the number of iterations. In each computation iteration, we load a column of non-zero kernel data into some SIMD-register, and then load into some other SIMD-registers the input data according to the corresponding column offset of the non-zero kernel data. After that, the loaded kernel and input data are multiplied and accumulated into intermediate results simultaneously with SIMD instructions.

In practice, this procedure is implemented with highly optimized assembly. The group size and the number of collectively computed output elements are determined by the bit-length and number of SIMD registers.

\subsection{Latency estimation model}
\begin{figure*}
  \centering
  \includegraphics[width=\linewidth]{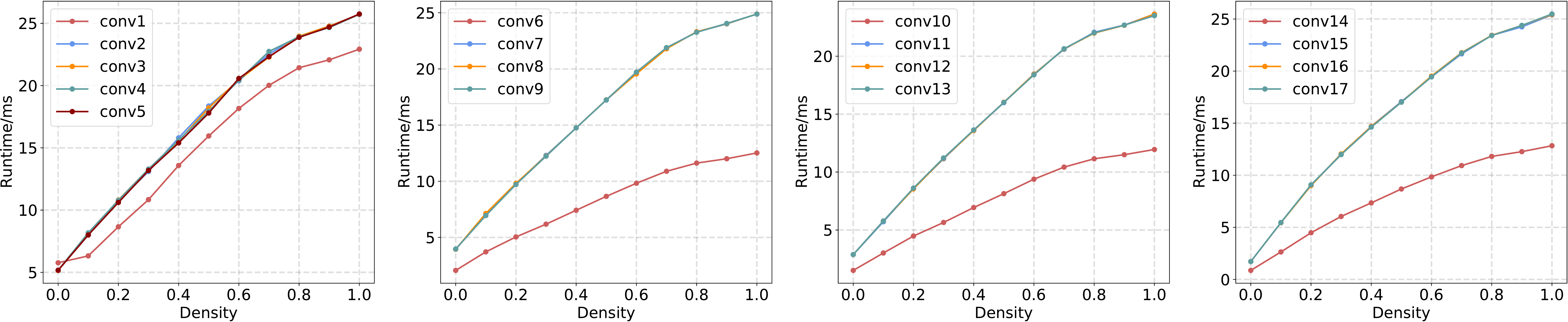}
  \caption{The density (ratio of non zero elements) of parameters \& the real runtime for each of convolutional layers in resnet18. Overall, the latency of each layer does not grow linearly with respect to the density, while the linearity is approximately satisfied locally. This motivates us to approximate the latency of each layer with linear interpolation.}
  \label{fig:time-ratio}
\end{figure*}
Having introduced the method utilized for CNN acceleration, the following problem is how to efficiently and accurately estimate the network latency given network parameters. 

The latency of the whole network can be denoted as the summation of the latency of each layer:
\begin{equation}
    \mathcal{T}(W) = \tau + \sum_{l=1}^L \mathcal{T}^{(l)}(W^{(l)})
    \label{eq:sum-layers}
\end{equation}
where $\mathcal{T}^{(l)}(W^{(l)})$ is the latency of the $l^{th}$ CONV/FC layer. $\tau$ denotes the latency of other layers (e.g. ReLU layers, pooling layers, et.al.) and can be regarded as a constant factor. In this work, the network is accelerated with sparsification, so the latency of the $l^{th}$ layer can be formulated as a function of the number of non-zero weights of its kernel tensor:
\begin{equation}
    \mathcal{T}^{(l)}(W^{(l)}) = \hat{\mathcal{T}}^{(l)}(\|W^{(l)}\|_0)
\end{equation}

Note that the above equation does not mean that the model size is used as a proxy of latency, because we model the latency layer-wisely, and the number of non-zero parameters across different layers may have different influences on the latency of the whole model. 

We propose to approximate $\hat{\mathcal{T}}^{(l)}(\cdot)$ with linear interpolation. This is based on the observation that the latency of each layer is locally linear with respect to the density of parameters as shown in Figure \ref{fig:time-ratio}. Take the $l^{th}$ layer as an example, we measure the run time of the layer on device when the number of non-zero parameters are $0=k_1^{(l)}< k_2^{(l)}\cdots < k_n^{(l)} = N^{(l)}$, respectively, where $N^{(l)}$ is the number of elements of $W^{(l)}$. For a given tensor $W^{(l)}$ with $s$ non-zero parameters, the run time can be then approximated by linear interpolation:
\begin{equation}
    \hat{\mathcal{T}}^{(l)}(s) = t_1^{(l)} + \sum_{i=1}^{n-1} \delta_i^{(l)}\alpha_i^{(l)}\min(k_{i+1}^{(l)} - k_i^{(l)}, s - k_i^{(l)})
\end{equation}
where $\delta_i^{(l)}$ is a variable which indicates whether the number of non-zero parameters is larger than $k_i^{(l)}$:
\begin{equation}
    \delta_i^{(l)} = \begin{cases}
    1, & s \ge k_i^{(l)} \\
    0, & otherwise
    \end{cases}
\end{equation}
and $t_i^{(l)}$ is the run time of the $l^{th}$ layer when the number of non-zero parameters is $k_i^{(l)}$. $\alpha_i^{(l)}$ is the ascending speed of latency when the number of non-zero parameters increases between $k_i^{(l)}$ and $k_{i+1}^{(l)}$:
\begin{equation}
    \alpha_i^{(l)} = \frac{t_{i+1}^{(l)} - t_{i}^{(l)}}{k_{i+1}^{(l)} - k_i^{(l)}}
    \label{eq:alpha-i}
\end{equation}

In practice, we set $k_1^{(l)}, k_2^{(l)} \cdots k_n^{(l)}$ to be $0, 0.1N^{(l)}, 0.2N^{(l)} \cdots N^{(l)}$. In this way, only 11 collective sparsity-latency data pairs are required for approximation of the network latency. We find that our approximation of latency with linear interpolation is rather accurate as shown in Figure \ref{fig:exp-linear-precision}. What's more, no platform knowledge is needed because we approximate the network latency with directly measurement and treat the hardware as a black-box. In contrast, previous works rely on either deep architecture knowledge \cite{EWP-2017,ECC-ICLR-2019} or a large collective~(usually over 10000) of sparsity-latency/energy data pairs for construction of look up table \cite{NET-ADAPT-2018} or training of the estimation model \cite{ECC-CVPR-2019}.

\subsection{The optimization algorithm}
\renewcommand{\algorithmicrequire}{\textbf{Input:}}
\renewcommand{\algorithmicensure}{\textbf{Output:}}
\begin{algorithm}[tb]
  \caption{Projection operation with bisection.}
  \label{alg:proj-bisection}
  \begin{algorithmic}[1]
    \REQUIRE {The variable $\tilde{U}$ to be projected. The group size $g$ for SIMD-structured pruning. The time budget $T_{budget}$. The tolerance $\epsilon$.}
    \ENSURE {The projected variable $U$.}
    \STATE {Divide $\tilde{U}$ into multiple groups with $g$ elements in each group. (Section \ref{sec:simd-structured-pruning})}
    \STATE {Sort the groups of elements in $\tilde{U}$ in term of $L_2$ norm.}
    \STATE {$N_{0}=0, N_{1}=$ the total number of groups in $\tilde{U}$.}
    \STATE {$t_{0}/t_{1}=$ the run-time of model if all the parameters are removed/retained.}
    \WHILE {$t_{1} - t_{0} > \epsilon$}
      \STATE {$N = \frac{N_{0}+N_{1}}{2}$}
      \STATE {$U=$ pick the top-N largest groups of elements in $\tilde{U}$}
      \STATE {$t=\mathcal{T}(U)$}
      \IF{$t<T_{budget}$}
        \STATE {$t_{0}=t, N_{0}=N$}
      \ELSE
        \STATE {$t_{1}=t, N_{1}=N$}
      \ENDIF
    \ENDWHILE
    \STATE {$U = $ pick the $\frac{N_{0}+N_{1}}{2}$ largest groups of elements in $\tilde{U}$.}
    \RETURN {$U$}
  \end{algorithmic}
\end{algorithm}
Now we have been able to efficiently approximate the latency of CNN models given parameters, we are ready to solve the constrained optimization problem as shown in Equation \ref{eq:problem-def}. Many optimization algorithms can be applied to solve the problem \ref{eq:problem-def}, such as Alternating Direction Method of Multipliers (ADMM) \cite{}, Projected Gradient Descent (PGD) \cite{}, and so on. In this paper, we apply the ADMM which is recently proved to be sufficient for solving non-convex, non-smooth problems \cite{ADMM-CONVERGENCE-2018}. One may choose other optimization algorithms to solve the problem \ref{eq:problem-def}, and the proposed SIMD-structured pruning and resource estimation method with linear interpolation are also applicable as a plugin for other network compression methods to further improve their latency-accuracy trade-off, while this is out of the scope of this paper. The original problem can be reformulated by:
\begin{equation}
  \min_{W,U}~~\mathcal{L}(W)~~s.t.~~W=U,\mathcal{T}(U)\le T_{budget}
\end{equation}
where $\mathcal{T}(\cdot)$ is defined by equations (\ref{eq:sum-layers})-(\ref{eq:alpha-i}). By applying augmented largrangian, the above problem is equivalent to:
\begin{equation}
\begin{split}
    \min_{W,U}\max_{Z}~~&\mathcal{L}(W)+<W-U,Z>+\frac{\rho}{2}\|W-U\|_2^2 \\
    s.t. ~~& \mathcal{T}(U)\le T_{budget}
\end{split}
\end{equation}
where $\rho > 0$ is a hyper-parameter. The main idea of ADMM is to update the original parameters $W$, the auxiliary variable $U$ and the dual variable $Z$ in an alternative manner:
\begin{subequations}
    \begin{align}
        W_{t+1} &= \arg\min_W~\mathcal{L}(W)+\frac{\rho}{2}\|W-U_t+\frac{Z_t}{\rho}\|^2 \label{eq:admm-update-w}\\
        U_{t+1} &=\arg\min_{\mathcal{T}(U)\le T_{budget}}~\|W_{t+1}-U+\frac{Z_t}{\rho}\|^2 \label{eq:admm-update-u}\\
        Z_{t+1} &= Z_t + \rho(W_{t+1}-U_{t+1})\label{eq:admm-update-z}
    \end{align}
\end{subequations}
The update of the original parameters $W$ and the dual variable $Z$ are relatively straight forward. For update of $W$, we apply SGD on the training dataset for one epoch. The main difficulty lies in the update of the auxiliary variable $U$, which is the projection of $\tilde{U}=W_{t+1}+\frac{Z_t}{\rho}$ on the constraint set. We solve this problem with a greedy algorithm. We first sort groups of elements of $\tilde{U}$ in term of $L_2$ norm, and pick them one by one, until the final latency achieves the target budget. Direct implementation of this algorithm is not efficient in that it may need a large number of iterations. While it can be efficiently implemented with bisection method as shown in Algorithm \ref{alg:proj-bisection}. After ADMM optimization finishes, we set $W = U$ and finetune the generated model on the training set for a few epochs. We summarize the final optimization algorithm in Algorithm \ref{alg:alcs}, and more details are given in Section \ref{sec:exp-setup}.

\begin{algorithm}
  \centering
  \caption{The ALCS algorithm}
  \label{alg:alcs}
  \begin{algorithmic}
    \REQUIRE{The base model with pretrained parameters $W$. The latency budget $T_{budget}$. The group size $g$ for SIMD-structured pruning. The budget tolerance $\epsilon$. The training dataset $\mathcal{D}$. The training epochs $E_{admm}$ and the penalty parameter $\rho$ for ADMM optimization. The training epochs $E_{ft}$ for finetuning.}
    \ENSURE{The pruned model with latency $T_{budget}$.}
    \STATE {Initialize $U_0 = \arg\min_{\mathcal{T}(U)\le T_{budget}}\|W-U\|^2$ with Algorithm \ref{alg:proj-bisection}, $Z_0 = W - U_0$}
    \FOR {$t = 1\rightarrow E_{admm}$}
      \STATE {Update $W$ with SGD on $\mathcal{D}$ for one epoch}
      \STATE {Update $U_t = \arg\min_{\mathcal{T}(U)\le T_{budget}}\|W - U + \frac{Z_{t-1}}{\rho}\|^2$ with Algorithm \ref{alg:proj-bisection}}
      \STATE {Update $Z_t = Z_{t-1} + \rho (W - U_t)$}
    \ENDFOR
    \STATE {Set $W = U_{E_{admm}}$}
    \STATE {Update the non-zero parameters of $W^*$ with SGD on $\mathcal{D}$ for $E_{ft}$ epochs}
    \RETURN{$W^*$}
  \end{algorithmic}
\end{algorithm}
\section{Experimental results}
\subsection{Experimental setup}
\label{sec:exp-setup}
We evaluate our method on both compact models such as MobileNet \cite{MBV1-2017}, as well as relatively heavy networks like ResNet18 and ResNet50 \cite{RESNET-2016} for 1000-class image classification on ImageNet \cite{IMGNET-2009}.  We do not conduct experiments on CIFAR because it is more practical and challenging to accelerate CNN models on large scale vision tasks. We use the standard data pre-processing pipeline which is provided by pytorch official examples. The batch size is set to 256, and the group size for SIMD-structured pruning is set to 4 to match the bit-length of SIMD registers
\footnote{In most mobile devices, length of each vector register for SIMD instructions is 128 bits, so for SIMD-structured pruning of 32-bit single float precision parameters, the group size should be 4.}. 
The hyper-parameter $\rho$ is set to 0.01 for all the experiments. In each ADMM iteration, for update of the original parameters $W$ as indicated in Equation (\ref{eq:admm-update-w}), we apply the momentum SGD optimizer for 1 epoch with learning rate fixed to 0.001 and weight decay set to $1e-4$ for ResNet and $4e-5$ for MobileNet. We apply 100 ADMM iterations for MobileNet and ResNet18 and 60 ADMM iterations for ResNet50. After ADMM iterations, the generated compressed model is then fine-tuned for 60 epochs with learning rate annealed from $0.001$ to $0$ with cosine learning rate scheduler. The weight decay is set to $0$. During this procedure, only non-zero parameters are updated.

The latency of all the dense models (including the models compressed with channel pruning methods) are measured with Tensorflow Lite \cite{TFLITE-2020}, which is one of the most popular mobile-oriented inference framework for DNNs, and the latency of all the SIMD-structured sparse models are measured with our proposed SIMD-structured sparse convolution computation algorithm, which is implemented in C++ with SIMD instructions. Averaged latency over 50 runs on a single ARM Cortex-A72 CPU is reported.

\subsection{Ablation study}
\paragraph{Precision of latency estimation:}
\begin{figure}[t]
    \centering
    \includegraphics[width=0.9\linewidth]{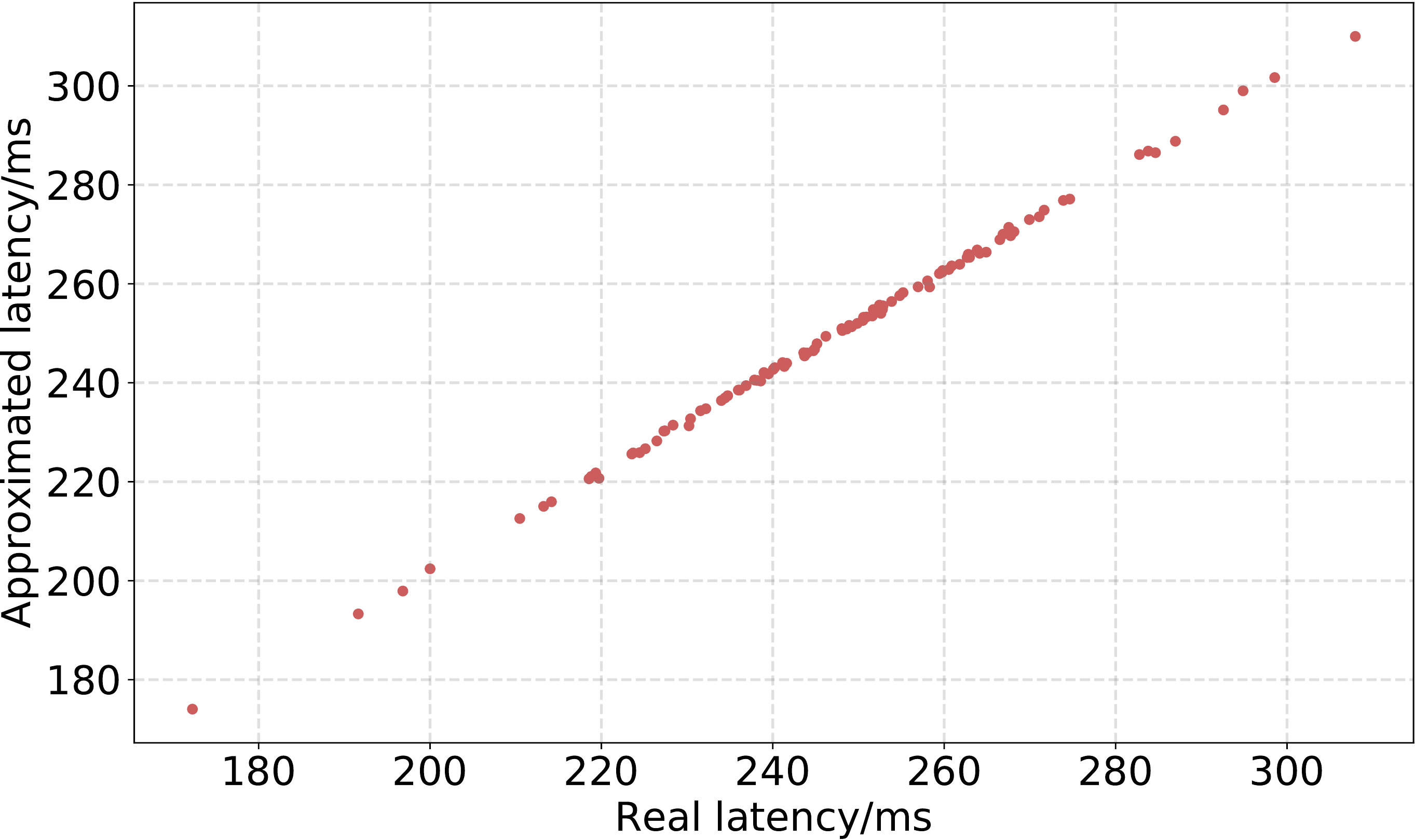}
    \caption{Real \& estimated latency on 100 randomly sampled ResNet18 models. Tested on a single ARM Cortex-A72 CPU.}
    \label{fig:exp-linear-precision}
\end{figure}
This section we first study the precision of the proposed latency estimation with linear interpolation. To this end, we uniformly sample 100 ResNet18 models with different sparsity, and plot the real latency and estimated latency in Figure \ref{fig:exp-linear-precision}. From the figure we can see that the proposed approximation of latency with linear interpolation is rather accurate.
\paragraph{Influence of \texorpdfstring{$\rho$}{Lg}:}
\begin{figure*}
    \centering
    \includegraphics[width=0.8\linewidth]{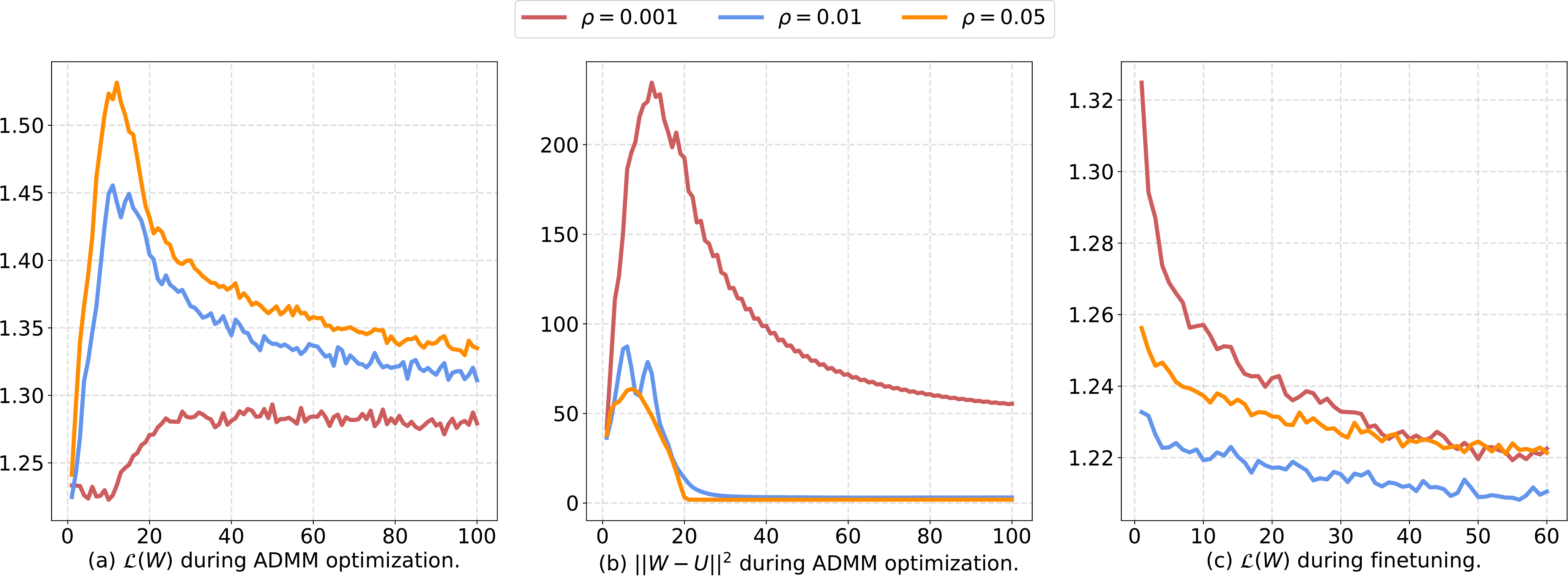}
    \caption{Training curves for compressing MobileNet on ImageNet during ADMM optimization and the following fine-tuning with different values of $\rho$. Better viewed in color.}
    \label{fig:exp-rho-influence}
\end{figure*}
\begin{table}
    \centering
    \begin{tabular}{l|ccc}
    \toprule
    \hline
    $\rho$ & FLOPs & Latency & Acc@1 \\
    \hline
    \tabincell{c}{0.001 \\ 0.01 \\ 0.05} & 185M & 62ms & \tabincell{c}{70.37\% \\ \textbf{70.53\%} \\ 70.38\%} \\
    \hline
    \bottomrule
    \end{tabular}
    \caption{Accuracy of compressed MobileNet on ImageNet with different values of $\rho$.}
    \label{tab:exp-rho-influence}
\end{table}

To study the influence of the hyper-parameter $\rho$ for our algorithm, we compress MobileNet on ImageNet and set the target latency to be 62ms with $\rho=\{0.001, 0.01, 0.05\}$. Results are shown in Figure \ref{fig:exp-rho-influence} and Table \ref{tab:exp-rho-influence}. From Figure \ref{fig:exp-rho-influence}(a) we can see that $\mathcal{L}(W)$ converges to a lower value with small $\rho$, since with smaller $\rho$, the algorithm focuses more on optimizing the original parameters $W$. While we can further see from Figure \ref{fig:exp-rho-influence}(b) that it is not sufficient to constraint the sparse structure of the original parameters $W$ if $\rho$ is too small. For instance, when $\rho=0.001$, the difference between $W$ and $U$ is rather large even after 100 ADMM iterations, which means that in this case $W$ fails to converge to be sparse during ADMM optimization. Therefore, during fine-tuning, the smallest $\rho$ doesn't lead to the lowest loss (see Figure \ref{fig:exp-rho-influence}(c)). From Table \ref{tab:exp-rho-influence}, we can see that $\rho=0.01$ achieves slightly better accuracy compared to the other two cases, and we will set $\rho=0.01$ in all the following experiments.

\paragraph{Impact of different components:}
\begin{table}
    \centering
    \begin{tabular}{l | c c c c c c}
       \toprule
       \hline
       \tabincell{c}{Pruning\\Method} & FLOPs & Latency & ADMM & FT & Acc@1 \\ \hline
       WP & 71.1M & 61.55ms & \checkmark & \checkmark & 68.36\% \\
       FP & 186M & 64ms & \checkmark & \checkmark & 67.76\% \\
       SIMD & 185M & 62ms & \checkmark & \checkmark & \textbf{70.53\%} \\
       SIMD & 185M & 62ms & & \checkmark & 70.08\% \\
       SIMD & 185M & 62ms & \checkmark & & 69.96\% \\
       \hline \bottomrule
    \end{tabular}
    \caption{Comparison of different variants of our method for compressing MobieNet on ImageNet. WP denotes weight pruning, FP denotes filter pruning, and SIMD denotes our proposed SIMD-structured pruning. ADMM and FT denotes the ADMM optimization process and the post fine-tuning process, respectively. Our method outperforms all the other variants in that it is able to achieve better trade-off between network latency and accuracy.}
    \label{tab:exp-simd-admm-ft}
\end{table}
\begin{figure*}
  \centering
  \includegraphics[width=0.8\linewidth]{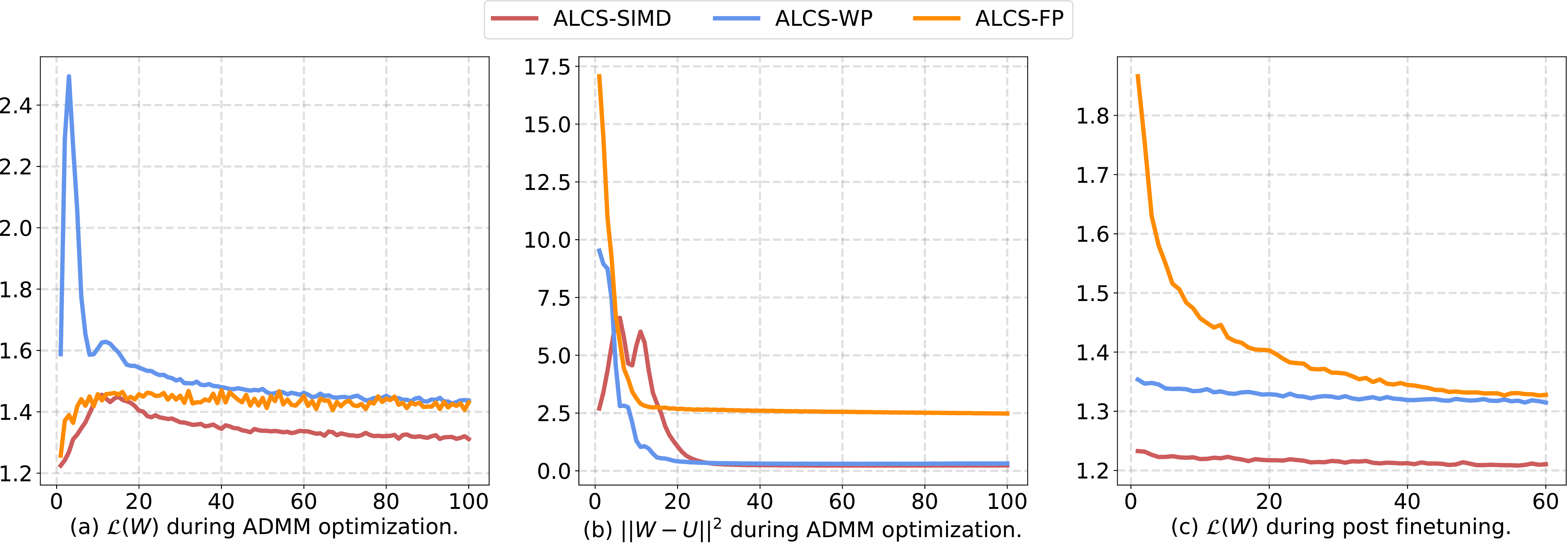}
  \caption{Training dynamics for compressing MobileNet on ImageNet during ADMM optimization and the post finetuning process with different pruning methods. SIMD denotes the proposed SIMD-structured pruning, WP denotes random weight pruning in which any individual element of parameters can be pruned without any constraint, and FP denotes filter pruning. The latency budget of all the methods are set to 62ms.}
  \label{fig:simd-wp-fp-training-dynamics}
\end{figure*}
\begin{figure}
    \centering
    \includegraphics[width=\linewidth]{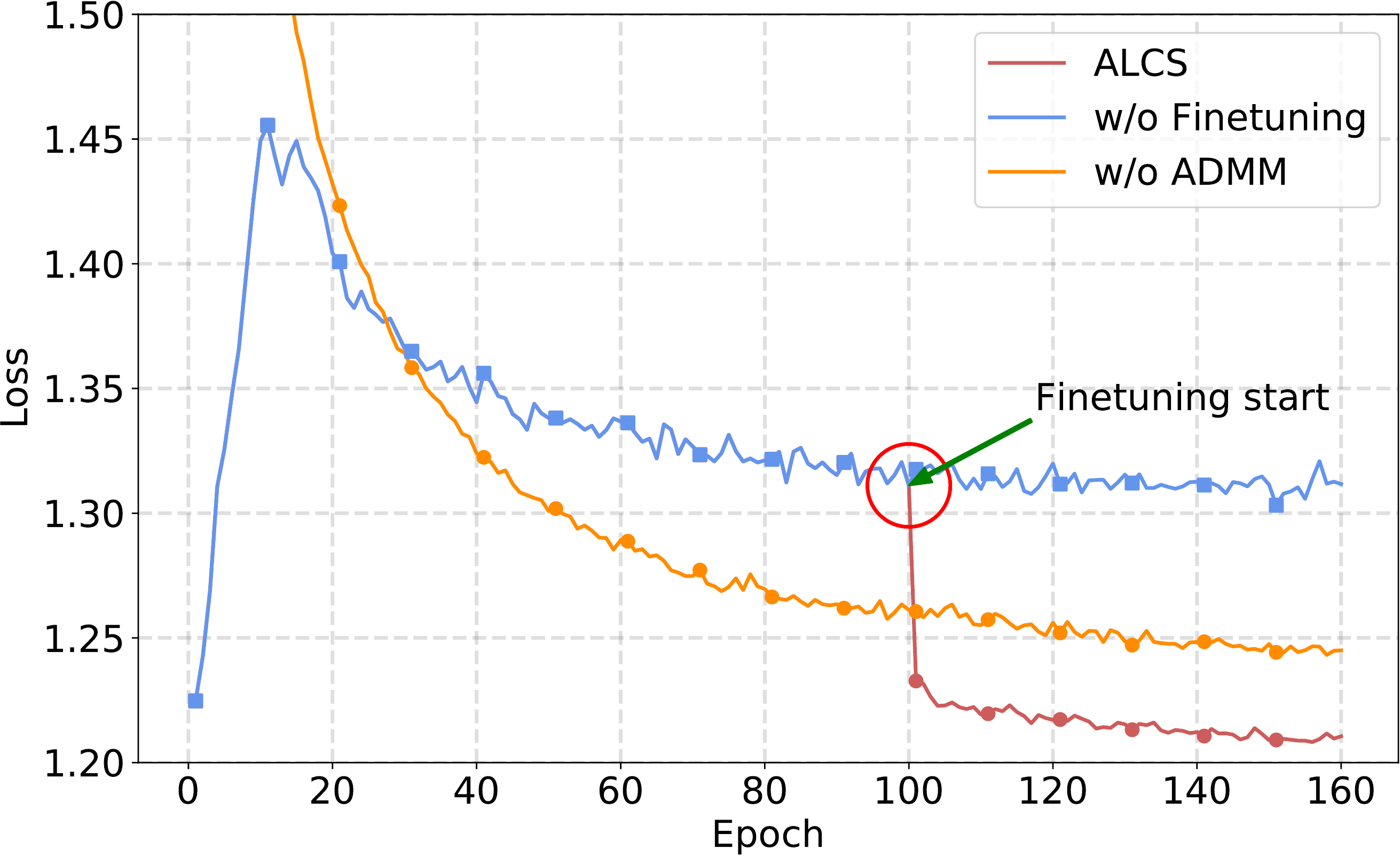}
    \caption{Comparison of different variants of ALCS with or without the ADMM optimization and the post finetuning process. The latency budget of all the compared methods are set to 62ms.}
    \label{fig:admm-ft}
\end{figure}
In this section, we study the impact of different components of ALCS. That is The SIMD-structured pruning, the ADMM optimization and the post finetuning. For this end, we compare several variants of ALCS for compressing MobileNet on ImageNet. Results are summarized in Table \ref{tab:exp-simd-admm-ft}, Figure \ref{fig:simd-wp-fp-training-dynamics} and Figure \ref{fig:admm-ft}. Where WP denotes weight pruning, in which each individual element of parameters can be removed or retained without any constraint. To measure the latency of models compressed by weight pruning, we apply the codes of XNNPACK \cite{XNNPACK-2020}, which is the state of the art implementation of sparse matrix multiplication \cite{FAST-2020}. Note that the implementation supports only matrix multiplication, which is equivalent with convolution layer with kernel size of $1\times 1$, so we do not prune the first convolution layer in this variant, following the same set as in \cite{FAST-2020}. FP denotes filter pruning, which prunes parameters in a channel wise manner. SIMD denotes the proposed SIMD-structured pruning. ADMM and FT denote the ADMM optimization and the post fine-tuning process, respectively. For fair comparison, we set the latency budgets to be the same and train all the variants with equal number of total epochs. Specifically, for ADMM+FT variants, we apply the same hyper-parameters as described in Section \ref{sec:exp-setup}. For ADMM-only variants, we apply 160 ADMM iterations, and for FT-only variants, we prune the model with norm-based method and employ fine-tuning for 160 epochs with learning rate fixed to 0.001 at the first 100 epochs and annealing from 0.001 to 0 with cosine learning rate at the last 60 epochs. 

As shown in table~\ref{tab:exp-simd-admm-ft}, the proposed training pipeline outperforms all the other variants. By comparing the first three variants, we can conclude that SIMD-structured pruning is able to achieve better trade-off between network accuracy and latency than weight pruning and filter pruning. For instance, the accuracy of ALCS with SIMD-structured pruning is $1.17\%$ higher than weight pruning and $1.77\%$ higher than filter pruning under similar latency budgets. This is mainly because that: (1) Compared to weight pruning, SIMD-structured pruning is more friendly to the SIMD architecture of mobile devices, and thus is able to achieve similar latency under a higher density, which benefits retaining accuracy of networks; (2) Compared to filter pruning, SIMD-structured pruning does not suffer from such strong constraint on data structure, thus improves flexibility and attains lower accuracy loss. 

By comparing the last three variants, we can see that both ADMM optimization and the post fine-tuning are necessary for improving the network accuracy. Particularly, when only applying the ADMM optimization, the final accuracy will be only $69.96\%$, which is $0.57\%$ lower than applying both ADMM optimization and the post finetuning, and the accuracy gap is $0.45\%$ without the ADMM optimization. In Figure \ref{fig:admm-ft} we further draw the training curves of these three variants, we see that there is a sharp decline in vallidation loss when the post finetuning begins. We explain that the ADMM optimization helps to find a better initialization for the post finetuning process.

\subsection{Comparison with state-of-the-arts}
\begin{table}[!bt]
    \centering
    \begin{tabular}{l|cccc}
         \toprule \hline
         Model & Method & FLOPs & Latency & Acc@1 \\ \hline
\multirow{3}{*}{ResNet18} & baseline & 1.8G & 537ms & 69.76\% \\
               & DMCP & 1.04G & 341ms & 69.70\% \\
               & ALCS(OURS) & 548M & 200ms & 69.88\% \\
               \hline
\multirow{9}{*}{ResNet50} & baseline & 4.1G & 1053ms & 76.15\% \\ \cline{2-5}
               & \multirow{2}{*}{DMCP} & 2.2G & 659ms & 76.2\% \\
               & & 1.1G & 371ms & 74.1\% \\ \cline{2-5}
               & HRank & 2.26G & 695ms & 75.56\% \\ \cline{2-5}
               & \multirow{3}{*}{AutoSlim} & 3.0G & 792ms & 76.0\% \\
               &                           & 2.0G & 609ms & 75.6\% \\
               &                           & 1.0G & 312ms & 74.0\% \\ \cline{2-5}
               & \multirow{2}{*}{ALCS(OURS)} & 2.2G & 630ms & 76.26\% \\
               & & 985M & 370ms & 75.05\% \\
               \hline
\multirow{13}{*}{MobileNet} & Uniform $1.0\times$ & 569M & 167ms & 71.8\% \\
               & Uniform $0.75\times$ & 325M & 102ms & 68.4\% \\
               & Uniform $0.5\times$ & 150M & 53ms & 64.4\% \\ \cline{2-5}
               & AMC & 285M & 94ms & 70.7\% \\
               & Fast$^*$ & 71.1M & 61ms & 68.4\% \\ \cline{2-5}
               & \multirow{2}{*}{AutoSlim} & 325M & 99ms & 71.5\% \\
               &                                                & 150M & 55ms & 67.9\% \\ \cline{2-5}
               & \multirow{2}{*}{USNet} & 325M & 102ms & 69.5\% \\
               &                                          & 150M & 53ms & 64.2\% \\ \cline{2-5}
               & \multirow{4}{*}{ALCS(OURS)} & 283M & 82ms & 72.04\% \\
               &  & 185M & 62ms & 70.53\% \\
               &  & 140M & 52ms & 69.16\% \\
               &  & 91M & 40ms & 65.96\% \\
               \hline
               \bottomrule
    \end{tabular}
    \caption{Comparison of ALCS with other state-of-the-art network compression/acceleration methods on ImageNet. We compare ALCS with the following 6 state-of-the-art network compression/acceleration methods: DMCP \protect\cite{DMCP-2020}, HRank \protect\cite{HRANK-2020}, Fast sparse convolution \protect\cite{FAST-2020}, AMC \protect\cite{AMC-2018}, AutoSlim \protect\cite{AUTOSLIM-2019}, and USNet \protect\cite{USNET-2019}. The method marked by $^*$ indicates that it is a weight pruning method, the latency of which is measured by XNNPACK \protect\cite{XNNPACK-2020}. For all the other baseline methods, we download the models released by the authors and measure their latency with TFLite \protect\cite{TFLITE-2020}.}
    \label{tab:exp-compare-sota}
\end{table}
\begin{figure}
    \centering
    \includegraphics[width=\linewidth]{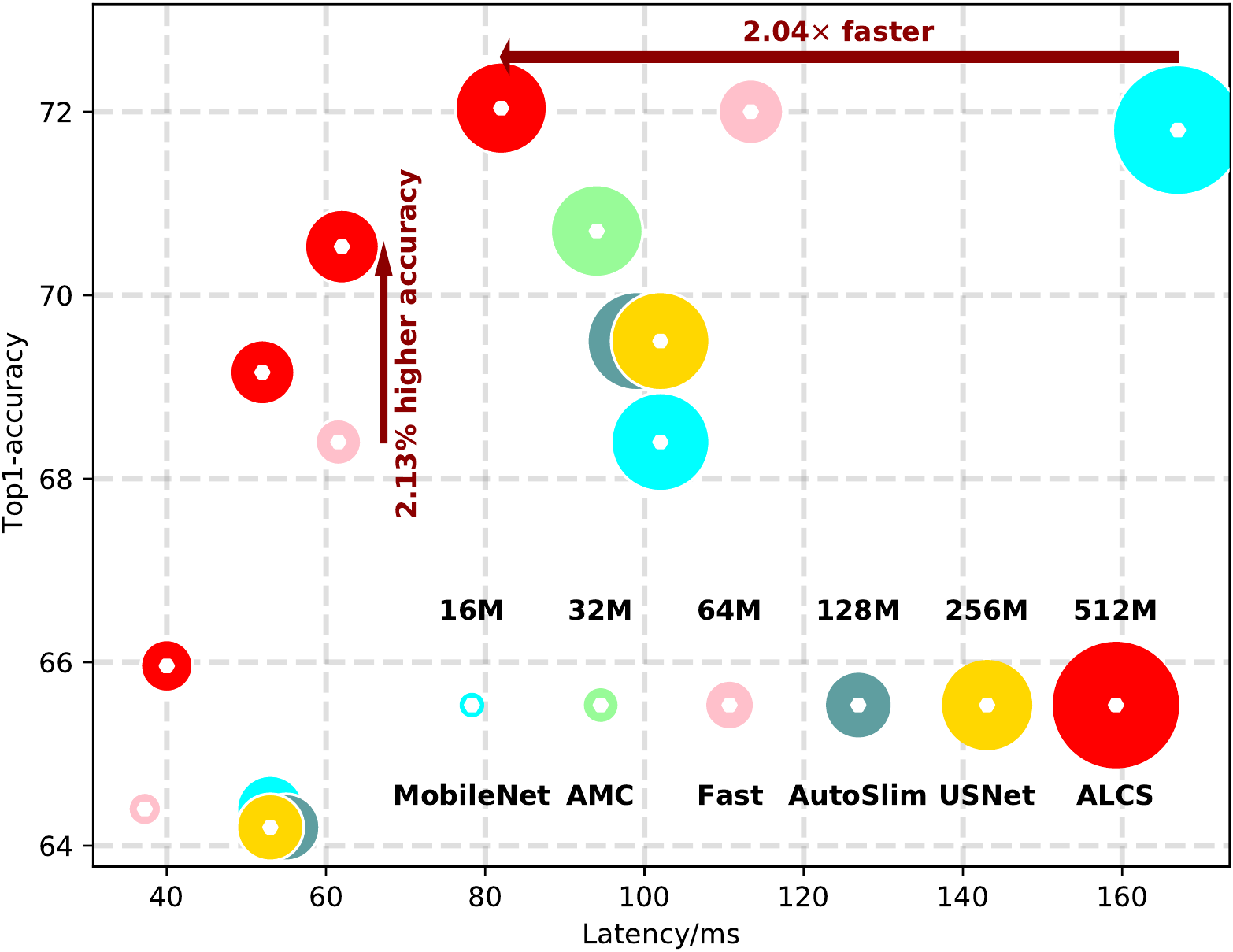}
    \caption{Comparison of ALCS with state of the art model acceleration methods for compressing MobileNet. The size of points denote the FLOPs of models. It is obviously that ALCS is able to achieve better accuracy-latency trade-off than all the other methods.}
    \label{fig:time-acc-flops}
\end{figure}
To further prove the efficacy of our method, in this section, we compare ALCS with various state-of-the-art network compression/acceleration methods. All the experiments are conducted on ResNet18, ResNet50 and MobileNet on ImageNet. For fair comparison, we set the latency budget to be the same for all approaches. The results are given in Table \ref{tab:exp-compare-sota} and Figure \ref{fig:time-acc-flops}.

From Table \ref{tab:exp-compare-sota}, we see that ALCS is able to achieve better trade-off between network accuracy and latency than all the other methods. For example, our method is able to accelerate the inference of ResNet18 by $2.68\times$ without any accuracy loss. Compared to DMCP \cite{DMCP-2020}, ALCS is $1.70\times$ faster with $0.18\%$ better accuracy. As for ResNet50, our method is $1.67\times$ faster with $0.11\%$ better accuracy and $2.85\times$ faster with only $1.1\%$ accuracy drop compared to the original model. On MobileNet, our method also achieves higher accuracy compared to other methods under similar or smaller latency budgets. For instance, under 82ms latency budget, ALCS achieves $72.04\%$ top-1 accuracy, which is $0.54\%$ higher than AutoSlim under the latency of 99ms and $1.34\%$ higher than AMC under the latency of 94ms. Compared to the original model, ALCS is $2.04\times$ faster without any accuracy loss. The same trend also holds under other latency budgets. Overall, the advantage of ALCS is more obvious on compact models under lower latency budgets. This implies that specialized design of pruning structure is more necessary for acceleration of compact models under tight latency budget. 

From Table \ref{tab:exp-compare-sota} and Figure \ref{fig:time-acc-flops} we see that ALCS does not achieve a better FLOPs-accuracy trade-off compared to Fast \cite{FAST-2020}. This is because that Fast accelerates the networks with random weight pruning, in which each individual element of parameters can be pruned without any constraint. In contrast, in ALCS, the proposed SIMD-structured pruning is used, in which a group of parameters (4 parameters in our experiments) must be pruned or retained simultaneously, so Fast is able to achieve higher accuracy than ALCS under the same FLOPs. Whereas the goal of this paper is not to reduce the model size or the number of arithmetic operations, but to accelerate the true inference speed, because when deploying deep models for practical applications, it is often the true runtime, instead of the FLOPs of models, that we concern more about. Compared to random weight pruning, the proposed SIMD-structured pruning fully utilizes the advantages of SIMD architectures in the target platform, which is helpful for achieving high computation efficiency. Thus, to achieve some latency budget, more parameters need to be pruned when using random weight pruning. For example, to accelerate the latency of MobileNet to $62ms$, the FLOPs of Fast is $71.1M$, which means that $\sim 90\%$ of parameters need to be pruned. On the contrary, the FLOPs of ALCS is $185M$, only $\sim 70\%$ of the parameters need to be pruned, which is conducive to enhance the model accuracy. As a result, ALCS is able to achieve better trade-off between accuracy and latency compared to Fast, as shown in Table \ref{tab:exp-compare-sota} and Figure \ref{fig:time-acc-flops}.
\section{Conclusion}
In this paper, we propose ALCS (Architecture Aware Latency Constrained Sparse Neural Networks) for model acceleration on mobile devices. Considering that most modern mobile devices utilize the Single Instruction Multiple Data (SIMD) technique to improve the computation capacity, we propose a novel SIMD-structured pruning method along with an efficient SIMD-structured sparse convolution algorithm for acceleration of sparse models. Moreover, we propose to estimate the latency of compressed models with piece wise linear interpolation, which is accurate and efficient, and does not need a large number of collective architecture-latency data pairs in comparison with existing budget approximation methods. The whole latency constrained problem is finally solved with ADMM. Extensive experimental results on various network architectures indicate that ALCS is able to achieve better latency-accuracy trade-off thanks to the proposed SIMD-structured pruning along with the efficient SIMD-structured sparse convolution algorithm.

The main purpose of this paper is to investigate the design space lying between the traditional random weight pruning and structured filter level pruning. The results show that it is possible to further push ahead the latency-accuracy frontier with the help of SIMD instructions in modern CPUs. One limitation of SIMD-structured pruning is that it is not applicable on GPUs because the computing architectures are very different, which is an interesting future direction of this work.

\ifCLASSOPTIONcaptionsoff
  \newpage
\fi



%


\bibliographystyle{IEEETran}
\bibliography{references}

\end{document}